\pdfoutput=1

\documentclass[11pt]{article}

\usepackage[final]{acl}

\usepackage{times}
\usepackage{latexsym}

\usepackage[T1]{fontenc}

\usepackage[utf8]{inputenc}

\usepackage{microtype}

\usepackage{inconsolata}

\usepackage{graphicx}

\usepackage{graphicx,xcolor}  
\usepackage{url}

\usepackage{multirow}
\usepackage{comment}
\usepackage{booktabs}
\usepackage[normalem]{ulem}
\usepackage[xcolor,framemethod=tikz]{mdframed}
\usepackage{collcell}
\usepackage{hhline}
\usepackage{colortbl}
\usepackage{adjustbox}
\usepackage{lingmacros}
\usepackage{subcaption}
\usepackage{mathtools}

\usepackage{amsmath} 
\usepackage{amssymb} 

\usepackage{tikz}
\usepackage{xspace}
\usepackage{fontawesome}

\usepackage[whole]{bxcjkjatype}

\def\colorModel{hsb}

\newcommand\ColCell[1]{
  \color{white} 
  \pgfmathsetmacro\compA{180/360}  
  \pgfmathsetmacro\compB{1}       
  \pgfmathsetmacro\compC{max(0, min(1, (#1-10)/90))}  
  \edef\x{\noexpand\centering\noexpand\cellcolor[\colorModel]{\compA,\compB,\compC}}\x #1
}

\newcommand\ColCellOrange[1]{
  \color{white} 
  \pgfmathsetmacro\compA{30/360}  
  \pgfmathsetmacro\compB{1}       
  \pgfmathsetmacro\compC{max(0, min(1, (#1-10)/90))}  
  \edef\x{\noexpand\centering\noexpand\cellcolor[\colorModel]{\compA,\compB,\compC}}\x #1
}

\newcolumntype{E}{>{\collectcell\ColCell}m{3.5ex}<{\endcollectcell}}

\newcolumntype{O}{>{\collectcell\ColCellOrange}m{3.5ex}<{\endcollectcell}}

\newcommand\items{14}   
\newcommand\rotation{30} 

\definecolor{backgray}{rgb}{0.9,0.9,0.9}

\newcommand{\un}{*}

\newlength{\astspace}
\newlength{\excspace}
\newcommand{\ac}{\hspace*{\astspace}}

\newcommand{\nextline}[1]{\\\hspace*{\astspace}#1}

\newcommand{\lone}{\ensuremath{\mathrm{L}_1}\xspace}
\newcommand{\ltwo}{\ensuremath{\mathrm{L}_2}\xspace}

\title{Developmentally-plausible Working Memory Shapes \\ a Critical Period for Language Acquisition}

\author{Masato Mita \and Ryo Yoshida \and Yohei Oseki \\
  The University of Tokyo \\
  \texttt{\{mita, yoshiryo0617, oseki\}@g.ecc.u-tokyo.ac.jp}}

\begin{document}
\maketitle
\begin{abstract}
Large language models possess general linguistic abilities but acquire language less efficiently than humans.
This study proposes a method for integrating the developmental characteristics of working memory during the critical period, a stage when human language acquisition is particularly efficient, into the training process of language models.
The proposed method introduces a mechanism that initially constrains~\textit{working memory} during the early stages of training and gradually relaxes this constraint in an exponential manner as learning progresses.
Targeted syntactic evaluation shows that the proposed method outperforms conventional methods without memory constraints or with static memory constraints.
These findings not only provide new directions for designing data-efficient language models but also offer indirect evidence supporting the role of the developmental characteristics of working memory as the underlying mechanism of the critical period in language acquisition.
\end{abstract}

\begin{center}
    \faGithub~\url{https://github.com/osekilab/CPLM}
\end{center}

\section{Introduction}
Large language models (LLMs) exhibit general linguistic abilities comparable to those of humans; however, their efficiency in language acquisition remains far inferior. 
It has been noted that LLMs require data quantities that are three to four orders of magnitude larger than those needed for humans to achieve comparable performance across many evaluation metrics~\cite{warstadt-etal-2023-findings}.
This disparity in data efficiency reflects the current reliance of LLMs on scaling and suggests not only a significant potential for improving learning efficiency but also the possibility of drawing \textit{insights} from human language processing and acquisition.

\begin{figure}[t]
 \centering
  \includegraphics[width=1.0\linewidth]{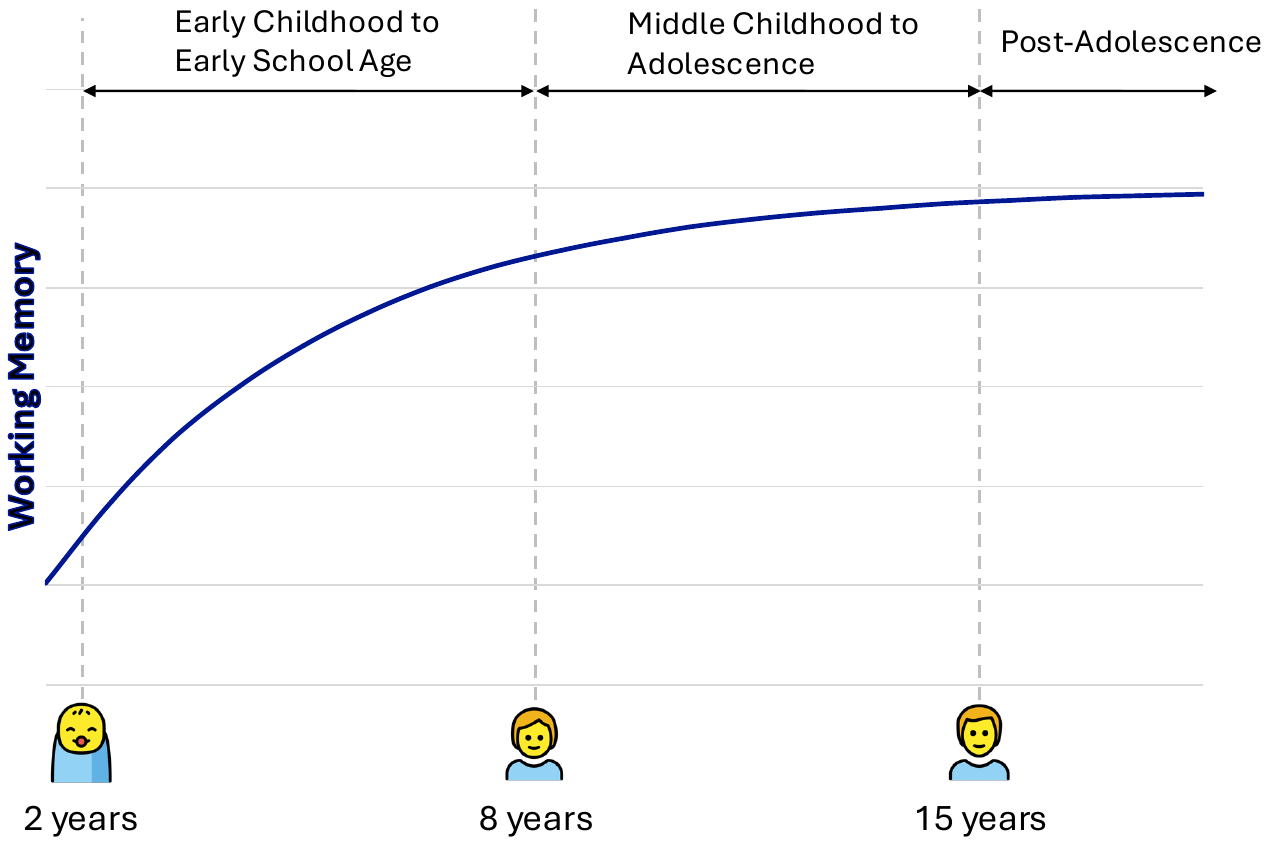}
 \caption{Developmental trajectory of human working memory}
 \label{fig:trajectory}
\end{figure}

An important theoretical framework for understanding the efficiency of human language acquisition is the \textbf{Critical Period Hypothesis (CPH)}~\cite{lenneberg1967biological}. 
The CPH posits that there is a specific period during which language can be acquired efficiently and that this ability diminishes thereafter. 
Various studies, including cases of limited exposure to first language (\lone) during childhood and age-related effects on second language (\ltwo) acquisition, support the existence of a critical period (CP)~\cite{FROMKIN197481,curtiss1977genie,JOHNSON198960}. 
However, the reasons why children acquire language more efficiently than adults remain partially unresolved.  
A compelling explanation is the \textbf{Less-is-More Hypothesis}~\cite{NEWPORT199011}, which attributes the superior learning of children to limited cognitive resources such as working memory.
According to this hypothesis, children's limited processing capacities enable them to efficiently extract fundamental patterns and structures (e.g., grammatical rules) from linguistic input, whereas adults, with their greater cognitive capacities, are more likely to be distracted by complex information, thereby hindering rule acquisition.

This hypothesis not only offers a compelling account of human learning but also has implications for how we design artificial systems. 
Language is not an arbitrary object of learning but a cultural artifact shaped under cognitive constraints. 
A growing body of work suggests that its structural properties reflect pressures for \textit{efficient communication}—that is, to maximize informational content while minimizing cognitive effort in production and comprehension under human limitations~\citep{zipf-1949,jaeger-tily-2011,christiansen-chater-2016,kemp-etal-2018,fedorenko-etal-2024}. 
Over generations, language has likely evolved to be learnable by agents with limited memory and processing capacity.
From this perspective, incorporating such constraints into language models (LMs) is not merely an act of mimicking human limitations, but a theoretically grounded way to introduce an inductive bias that aligns with the nature of the target: language shaped by cognitively bounded agents. 
Learning under such constraints may help LMs acquire representations better suited to natural language.\footnote {\citet{futrell2025linguisticslearnedstopworrying} for an alternative view suggesting that, given the empirical success of machine learning, effective learning may not require cognitively inspired constraints or inductive biases.}

Inspired by the \textit{Less-is-More} hypothesis, we use LMs to study the CP for language acquisition, focusing on \lone acquisition and investigating whether integrating human cognitive developmental characteristics, particularly the developmental properties of \textit{working memory} (Figure~\ref{fig:trajectory}), into LMs can facilitate efficient language acquisition. 
Specifically, we propose a method for incorporating the exponential increase in working memory capacity that corresponds to the CP into LMs and analyze its impact on learning efficiency. 
Using a GPT-2 model~\cite{radford2019language} trained on a Child-Directed Speech (CDS) dataset~\cite{aochildes2021}, we conduct evaluation experiments with Zorro~\cite{huebner-etal-2021-babyberta}, a targeted syntactic evaluation benchmark specialized for CDS. 
The results demonstrate that a cognitively plausible model, which initially restricts working memory and gradually relaxes this constraint exponentially as training progresses, outperforms models without memory constraints or with static memory constraints.
These findings provide new insights into designing data-efficient LMs, contributing to the field of \textbf{natural language processing}, while also offering indirect evidence supporting the role of the developmental characteristics of working memory as the underlying mechanism of the CPH in human language acquisition, contributing to the field of \textbf{cognitive science}.

\section{Related Work}
\subsection{Critical Period for Language Acquisition}
\label{subsec:cp}

The CPH posits that language acquisition is most efficient within a specific developmental window, after which it declines. 
CP effects are observed in both \lone and \ltwo acquisition, suggesting a shared underlying mechanism.

\paragraph{Critical Period for \lone Acquisition}
Research in neurolinguistics and cognitive science suggests that there is a biologically determined CP for acquiring an~\lone, beyond which full native-like proficiency is unattainable if exposure to language is delayed. 
Studies on late \lone learners, such as deaf individuals who acquire sign language after early childhood, indicate severe deficits in grammatical proficiency compared to those exposed to language from birth~\cite{mayberry1989looking, NEWPORT199011}. 
These findings suggest that neural plasticity, essential for \lone acquisition, diminishes with age, limiting the ability to develop full linguistic competence.
From a theoretical perspective, the existence of the CP for \lone acquisition is often attributed to biological constraints. 
Nativist theories propose that \lone acquisition relies on an innate language faculty that operates most effectively during the CP~\cite{penfield1965conditioning,chomsky1965,pinker1994language}. 
On the other hand, empiricist perspectives argue that the decline in \lone learning ability may result from environmental factors, such as a reduced need for language learning mechanisms once fundamental linguistic structures have been internalized~\cite{elman1996rethinking,seidenberg2006connectionist}. 
Despite extensive research, the precise boundary and mechanisms of the CP for \lone remain a subject of debate.

\paragraph{Critical Period for \ltwo Acquisition}
CP effects are also observed in \ltwo acquisition, where late learners struggle with pronunciation, morphology, and syntax~\cite{JOHNSON198960,hartshorne2018critical}. 
While biological constraints play a role, entrenchment—where prior exposure to \lone limits flexibility in learning new linguistic structures—is also a factor~\cite{ellis2000age, seidenberg2006connectionist}. 
Although the CP for \ltwo acquisition is an important topic, this study focuses on the CP for \lone acquisition, since our goal is to design data-efficient LMs by exploring the mechanisms of CP in \lone acquisition.

\subsection{The Role of Language Models in Acquisition Theories}

In recent years, computational models have played a crucial role in elucidating the mechanisms of language acquisition. 
These models enable controlled investigations of learning mechanisms and environments, which are difficult to achieve with human participants, and they are used to test theoretical claims such as the ``poverty of the stimulus''~\cite{Clark-etal-2011}. 
For instance, \citet{McCoy-etal-2020}, \citet{Wilcox-etal-2024}, and~\citet{warstadt-etal-2023-findings} have employed LMs to directly test hypotheses about language acquisition, demonstrating that such models can provide proof-of-concept evidence for \textit{learnability}. 
These studies have attracted attention as efforts to deepen theoretical discussions on language acquisition through computational modeling.

While Transformer-based LMs are not cognitive models in a strict sense, they are widely adopted in acquisition research as abstract ``model learners''~\cite{warstadt2022artificial}. 
Rather than replicating the full complexity of human cognition, these models are used to investigate the role of specific biases by selectively adding or removing them. 
This approach allows researchers to assess whether certain linguistic phenomena can be acquired purely through statistical learning or require inductive constraints, thereby testing the necessity of such biases. 
When a model fails to acquire a phenomenon in the absence of a particular bias, but succeeds once the bias is introduced, it offers at least weak evidence for the bias's relevance in human language acquisition~\cite{McCoy-etal-2020}. 
Our study follows this reverse-engineering paradigm~\cite{DUPOUX201843}, using LMs not as literal simulations of human learners, but as controlled testbeds for cognitive hypotheses.

\citet{Constantinescu-tacl2025} investigated CP phenomena in \ltwo acquisition and \lone attrition,\footnote{The phenomenon in which earlier cessation of \lone exposure increases the likelihood of \lone forgetting.} assuming a shared underlying mechanism for CP effects across \lone and \ltwo. 
They simulated \ltwo exposure at varying ages to examine how LMs differ from human learners, finding that LMs do not naturally exhibit CP effects. 
To artificially induce such effects, they employed Elastic Weight Consolidation~\cite{Kirkpatrick-etal-2017}, a regularization method for mitigating catastrophic forgetting, thereby mimicking a maturational decline in plasticity. 
Their findings suggest that CP effects are not an inevitable outcome of statistical learning but may instead involve innate mechanisms.
While this study shares the broader objective of enhancing the cognitive plausibility of LMs as models of human language acquisition, it differs from \citet{Constantinescu-tacl2025} in both \textit{focus} and \textit{methodology}. 
Rather than modeling CP effects through dataset manipulation or post-CP plasticity constraints, this study explicitly addresses the \textbf{developmental processes unfolding during the CP itself}. 
Specifically, we integrate a mechanism to simulate the progressive growth of working memory capacity throughout the CP, a factor considered crucial for \lone acquisition but previously unmodeled in LM-based research. 
By incorporating developmental constraints, this study aims to provide a more fine-grained computational model of early \lone acquisition and its cognitive underpinnings, advancing the developmental plausibility of LMs.

\section{Critical Period-inspired Language Model}
\label{sec:proposed_method}

\subsection{Modeling Developmental Trajectory of Human Working Memory}
\label{subsec:wm_trajectory}
Human working memory undergoes substantial developmental changes, progressing through three distinct stages: early childhood to early school age (2–7 years), middle childhood to early adolescence (8–14 years), and post-adolescence (15 years and older). 
During early childhood, both information retention capacity and processing ability improve rapidly, reflecting a significant expansion of cognitive resources~\cite{Cowan1999TheRO, Gathercole-etal-2004}. 
This rapid growth begins to decelerate during middle childhood and early adolescence as the brain approaches maturation~\cite{Luna-etal-2004, Gathercole-etal-2004}. 
By post-adolescence, working memory capacity plateaus, reaching adult-level performance~\cite{Sowell2002DevelopmentOC, Luna-etal-2004}.

Based on these observations, we characterized the growth trajectory of working memory, as illustrated in Figure~\ref{fig:trajectory}, using an exponential model of the form \( y = b - a^x \) (\( 0 < a < 1 \)).
In this model, \( b \) represents the asymptotic upper limit of working memory capacity, corresponding to adult-level performance, while \( a \) determines the rate of growth. 
Specifically, smaller values of \( a \) result in steeper early growth, reflecting the rapid cognitive development observed during early childhood, whereas larger values of \( a \) indicate a slower rate of change. 

This modeling approach is justified for several reasons. 
First, the horizontal asymptote inherent in the exponential function accurately represents the biological ceiling of adult working memory capacity.
Second, the rapid initial increase observed during early childhood is consistent with the steep growth predicted by this exponential form. 
Finally, alternative models, such as logarithmic or linear growth, fail to account for both the early rapid development and the eventual plateau: logarithmic models imply unbounded growth, while linear models oversimplify the deceleration phase.
Thus, the exponential model \( y = b - a^x \) offers a concise and biologically plausible representation of the developmental trajectory of human working memory, aligning well with observed patterns and theoretical considerations.\footnote{ACT-R~\cite{Anderson1989-ANDHMA} suggests that working memory \textit{decays} exponentially in language processing, while we propose that working memory \textit{grows} exponentially in language acquisition, but whether the shared exponential function between language processing and acquisition is a coincidence remains to be investigated in future.}

\subsection{Integrating Human Working Memory into Language Models}
 
In this study, Attention with Linear Biases (ALiBi)~\cite{press2022train} is employed to model the constraints of human working memory.   
ALiBi is a method for Transformer~\cite{NIPS2017_3f5ee243} models that does not use positional embeddings but instead applies a distance-dependent linear penalty to attention scores.  
Specifically, the attention score for an input sequence of length \(L\) is calculated as follows:

\begin{equation}
\begin{aligned}
\text{Attention Score} &= \text{softmax}\left(q_i K^\top + m \cdot B \right), \\
B &= \begin{bmatrix}
-(i-1) & -(i-2) & \cdots & 0
\end{bmatrix}.
\end{aligned}
\label{eq:attention}
\end{equation}

Here, \( q_i \in \mathbb{R}^{1 \times d} \), \( K \in \mathbb{R}^{L \times d} \), \( m \in \mathbb{R}_{[0,1]} \), and \( B \in \mathbb{R}^{1 \times L} \) represent the query, the key, a scalar slope specific to each attention head, and a bias matrix encoding the relative distances between queries and keys, respectively, where \( B_{i} \) is defined as the negative absolute difference between the query position \( i \) and each key position.
The values of \(m\) are set geometrically for each head. 
For example, in an 8-head model, the values of \(m\) are assigned as follows: \(m = 1, \frac{1}{2}, \frac{1}{4}, \ldots, \frac{1}{128}\). 
The slope \(m\) takes values in the range \([0, 1]\), ensuring a consistent interpretation of its influence on attention scores.
By penalizing attention scores for query-key pairs with greater distances, ALiBi introduces a \textit{recency bias} to the model. 
Originally, ALiBi was proposed to enhance the extrapolation capability of Transformer models.
More recently, \citet{clark-etal-2025-linear} has shown that incorporating it into attention score computation during training allows for the estimation of surprisal patterns resembling human reading times.  
This suggests its potential for modeling human-like memory decay and cognitive limitations.

However, since the slope \(m\) in ALiBi is fixed for each attention head, the approach does not inherently reflect the developmental increase in working memory capacity (i.e., reduced decay) over time (Figure~\ref{fig:trajectory}). 
Therefore, this study proposes a method, \textbf{\textsc{DynamicLimit-Exp}}, which replicates the developmental characteristics of working memory during the CP, specifically its exponential growth. 
This is achieved by exponentially decreasing the slope \( m \) in ALiBi as training epochs progress.  
In this method, the slope \(m\) in the ALiBi mechanism is updated at each epoch \(t\) as follows:

\begin{equation}
m_t = m_0 \cdot r^t,
\end{equation}
where \(m_0\) represents the initial slope, \(r \in (0, 1)\) is the decay rate, and \(t\) denotes the current epoch. 
In this study, the model's working memory capacity \(w_t\) is formulated as follows:

\begin{equation}
w_t \coloneqq 1 - m_t.
\end{equation}

This definition links the dynamically decaying slope \(m_t\) to the model's working memory capacity \(w_t\): as \(m_t\) decreases exponentially, \(w_t\) grows, enabling broader contextual retention over time. 
By simulating this developmental trajectory, the model initially focuses on short-range dependencies and gradually attends to longer ones.

\begin{figure}[t]
 \centering
  \includegraphics[width=1.0\linewidth]{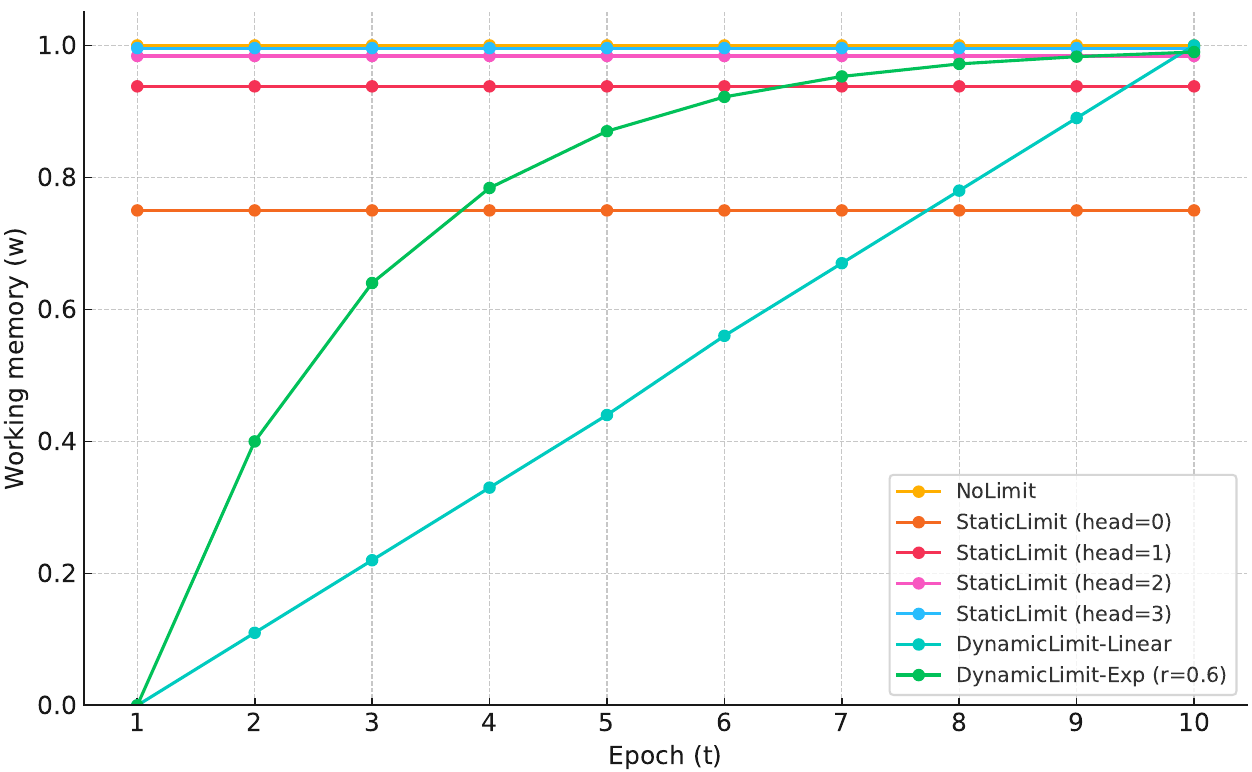}
 \caption{Trajectory of working memory capacity for each model (num. of epochs = 10)}
 \label{fig:wm_curves}
\end{figure}

\section{Experiments}
\label{sec:experiment}
This study explores whether LMs trained from scratch can achieve more efficient \lone acquisition by incorporating the developmental characteristics of human working memory.
Specifically, we aim to determine whether this approach can replicate the increased efficiency of \lone acquisition observed during the CP in \lone acquisition, focusing on the developmental advantages before the end of this period.

\subsection{Configurations}
\paragraph{Models}

\begin{table*}[t!]\small\centering
\begin{adjustbox}{max width=\textwidth}
\renewcommand{\arraystretch}{1.1}
\begin{tabular}{l*{\items}{|E}|}
\multicolumn{1}{p{2.5ex}}{\rotatebox{0}{Model}} &
\multicolumn{1}{p{2.5ex}}{\rotatebox{\rotation}{\textbf{\textsc{Overall}$^{*}$}}} &
\multicolumn{1}{p{2.5ex}}{\rotatebox{\rotation}{\textsc{D-n agr}}} &
\multicolumn{1}{p{2.5ex}}{\rotatebox{\rotation}{\textsc{S-v agr}}} &
\multicolumn{1}{p{2.5ex}}{\rotatebox{\rotation}{\textsc{Ana.\ agr}}} &
\multicolumn{1}{p{2.5ex}}{\rotatebox{\rotation}{\textsc{Arg.\ str}$^{*}$}} &
\multicolumn{1}{p{2.5ex}}{\rotatebox{\rotation}{\textsc{Binding}$^{\dagger}$}} &
\multicolumn{1}{p{2.5ex}}{\rotatebox{\rotation}{\textsc{Case}$^{*}$}} &
\multicolumn{1}{p{2.5ex}}{\rotatebox{\rotation}{\textsc{Ellipsis}$^{\dagger}$}} &
\multicolumn{1}{p{2.5ex}}{\rotatebox{\rotation}{\textsc{Filler.\ gap}$^{*}$}} &
\multicolumn{1}{p{2.5ex}}{\rotatebox{\rotation}{\textsc{Irregular}}} &
\multicolumn{1}{p{2.5ex}}{\rotatebox{\rotation}{\textsc{Island}$^{\dagger}$}} &
\multicolumn{1}{p{2.5ex}}{\rotatebox{\rotation}{\textsc{Local.\ atr}$^{*}$}} &
\multicolumn{1}{p{2.5ex}}{\rotatebox{\rotation}{\textsc{Quantifiers}$^{*}$}}  & 
\multicolumn{1}{p{2.5ex}}{\rotatebox{\rotation}{\textsc{NPI}$^{*}$}} \\ \hhline{~*\items{|-}|}
\textsc{NoLimit} & 56.5 & 49.8 & 49.7 & 49.9 & 44.8 & 61.8 & 70.8 & 73.3 & 72.1 & 51.7 & 61.7 & 47.1 & 47.9 & 53.9 \\ \hhline{~*{14}{|-}|}
\textsc{StaticLimit} & 56.8 & 50.2 & 49.9 & 49.8 & 44.4 & 60.5 & 70.3 & 71.4 & 74.7 & 52.2 & 62.9 & 45.3 & 52.3 & 54.4 \\ \hhline{~*{14}{|-}|}
\textsc{DynamicLimit-Linear} & 61.6 & 51.0 & 49.6 & 49.5 & 64.3 & 60.3 & 88.6 & 47.6 & 90.8 & 53.0 & 57.0 & 47.9 & 56.8 & 84.3 \\ \hhline{~*{14}{|-}|}
\textsc{DynamicLimit-Exp} & 62.2 & 50.8 & 50.0 & 49.6 & 67.7 & 58.7 & 95.2 & 43.1 & 93.6 & 52.2 & 53.6 & 51.3 & 57.6 & 85.0 \\ 
\end{tabular}
\end{adjustbox}
\caption{
Accuracy (\%) of models trained on AO-CHILDES dataset. 
$^{*}$ and $^{\dagger}$ indicate items where \textsc{DynamicLimit-Exp} performed significantly better or worse than \textsc{NoLimit}, respectively (z-test for proportions, p~<~0.05).
}
\label{tab:main_result}
\end{table*}

\begin{table*}[t!]\small\centering
\begin{adjustbox}{max width=\textwidth}
\renewcommand{\arraystretch}{1.1}
\begin{tabular}{l*{\items}{|O}|}
\multicolumn{1}{p{2.5ex}}{\rotatebox{0}{Model}} &
\multicolumn{1}{p{2.5ex}}{\rotatebox{\rotation}{\textbf{\textsc{Overall}$^{*}$}}} &
\multicolumn{1}{p{2.5ex}}{\rotatebox{\rotation}{\textsc{D-n agr}}} &
\multicolumn{1}{p{2.5ex}}{\rotatebox{\rotation}{\textsc{S-v agr}}} &
\multicolumn{1}{p{2.5ex}}{\rotatebox{\rotation}{\textsc{Ana.\ agr}}} &
\multicolumn{1}{p{2.5ex}}{\rotatebox{\rotation}{\textsc{Arg.\ str}$^{*}$}} &
\multicolumn{1}{p{2.5ex}}{\rotatebox{\rotation}{\textsc{Binding}$^{\dagger}$}} &
\multicolumn{1}{p{2.5ex}}{\rotatebox{\rotation}{\textsc{Case}$^{*}$}} &
\multicolumn{1}{p{2.5ex}}{\rotatebox{\rotation}{\textsc{Ellipsis}$^{\dagger}$}} &
\multicolumn{1}{p{2.5ex}}{\rotatebox{\rotation}{\textsc{Filler.\ gap}$^{*}$}} &
\multicolumn{1}{p{2.5ex}}{\rotatebox{\rotation}{\textsc{Irregular}}} &
\multicolumn{1}{p{2.5ex}}{\rotatebox{\rotation}{\textsc{Island}$^{*}$}} &
\multicolumn{1}{p{2.5ex}}{\rotatebox{\rotation}{\textsc{Local.\ atr}$^{*}$}} &
\multicolumn{1}{p{2.5ex}}{\rotatebox{\rotation}{\textsc{Quantifiers}}}  & 
\multicolumn{1}{p{2.5ex}}{\rotatebox{\rotation}{\textsc{NPI}$^{*}$}} \\ \hhline{~*\items{|-}|}
\textsc{NoLimit} & 54.7 & 50.3 & 50.0 & 47.2 & 68.4 & 62.6 & 73.4 & 60.8 & 42.9 & 53.4 & 51.1 & 42.7 & 41.2 & 42.6 \\ \hhline{~*{14}{|-}|}
\textsc{StaticLimit} & 54.7 & 50.4 & 50.0 & 47.1 & 73.7 & 61.2 & 87.4 & 57.3 & 56.1 & 52.3 & 53.0 & 40.8 & 42.0 & 38.9 \\ \hhline{~*{14}{|-}|}
\textsc{DynamicLimit-Linear} & 58.6 & 50.0 & 50.5 & 48.4 & 71.9 & 58.8 & 96.9 & 38.7 & 82.7 & 51.6 & 57.9 & 59.6 & 41.5 & 53.4 \\ \hhline{~*{14}{|-}|}
\textsc{DynamicLimit-Exp} & 59.1 & 49.8 & 50.4 & 46.0 & 71.5 & 59.3 & 97.7 & 37.4 & 86.5 & 51.1 & 58.0 & 60.5 & 42.2 & 53.9 \\ 
\end{tabular}
\end{adjustbox}
\caption{
Accuracy (\%) of models trained on Wikipedia dataset. 
$^{*}$ and $^{\dagger}$ indicate items where \textsc{DynamicLimit-Exp} performed significantly better or worse than \textsc{NoLimit}, respectively (z-test for proportions, p~<~0.05).
}
\label{tab:wiki_result}
\end{table*}

We used the \texttt{transformers}~\cite{wolf-etal-2020-transformers} implementation of the GPT-2~\cite{radford2019language} as the base LM. 
While some studies utilize RoBERTa~\cite{liu2019robertarobustlyoptimizedbert} as a base model~\cite{huebner-etal-2021-babyberta, warstadt-etal-2023-findings}, we selected GPT-2 for two primary reasons: (1) its unidirectional (left-to-right) predictions more effectively capture human working memory constraints, and (2) GPT-based architectures dominate modern LLMs~\cite{openai2023gpt4,touvron2023llama}.

\paragraph{Dataset}

We used AO-CHILDES~\cite{aochildes2021}\footnote{\url{https://github.com/UIUCLearningLanguageLab/AOCHILDES}} as the training dataset, which is derived from the CHILDES dataset~\cite{Macwhinney2000} and records CDS from conversations between children and adults.
AO-CHILDES contains 5 million words of speech directed at English-speaking children aged 1–6 years and controls for external factors such as age group, speaker variation, and situational context.
As a preprocessing step, following \citet{haga-etal-2024-modeling}, all sentences were converted to lowercase, and sentences shorter than three words were excluded.
Since the AO-CHILDES dataset contains only about 5 million words, training a standard GPT-2 model would likely result in overfitting.  
To mitigate this, we followed existing studies on small language models (SLMs) trained with CDS datasets~\cite{huebner-etal-2021-babyberta,haga-etal-2024-modeling} and constructed an SLM with 4 layers, 4 attention heads, and 256 embedding dimensions for the base model.  
Details of the training configuration for the base model are provided in Appendix~\ref{appendix:detailed_settings}.

Furthermore, to determine whether the CP effect stems from exposure to specific linguistic stimuli, such as CDS, or from the model's cognitive developmental properties independent of input, we conducted a complementary experiment using Wikipedia (written language, adult-oriented) as training data.  
Following~\citet{huebner-etal-2021-babyberta}, 500,000 sentences were randomly sampled from the English Wikipedia corpus.
We used the latest version of Wikipedia, as of January 2025,\footnote{\url{https://dumps.wikimedia.org/enwiki/latest/enwiki-latest-pages-articles.xml.bz2}} and preprocessed it using  \texttt{WikiExtractor}.\footnote{\url{https://github.com/attardi/wikiextractor}}
We provide the sentence length distribution for the AO-CHILDES and Wikipedia datasets used in this experiment in Appendix~\ref{appendix:dist}.

\paragraph{Evaluation}

We evaluate the grammatical abilities of these models using a developmentally inspired targeted syntactic evaluation benchmark, Zorro~\cite{huebner-etal-2021-babyberta}.
Zorro is designed for assessing the syntactic and grammatical knowledge of LMs in child-directed language and consists of 13 mid-level categories and 23 subcategories.  
Each subcategory contains 2,000 sentence pairs, with one grammatically acceptable and one unacceptable sentence per pair.\footnote{While Zorro lacks naturalness, this can aid in isolating syntactic ability from lexical or semantic cues~\cite{gulordava-etal-2018-colorless}.}
Below is an example of a minimal pair from the ``Subject-verb agreement (\textsc{S-v agr})'' category:\footnote{See Appendix~\ref{appendix:zorro_items} for the full list of grammatical categories.}

\eenumsentence[1]{
    \item \ac The \textbf{lie} on the foot is flat.
    \item \un The \textbf{lies} on the foot is flat.
}

By inputting both the acceptable and unacceptable sentences into the model and calculating the proportion of pairs where the model assigns a higher probability to the acceptable sentence, we obtain the grammaticality judgment score (Accuracy).
In this study, we report scores for each mid-level category (henceforth, \textit{grammatical items}) as well as their macro-average.

\begin{table*}[t!]\small\centering
\begin{adjustbox}{max width=\textwidth}
\renewcommand{\arraystretch}{1.1}
\begin{tabular}{lrrrrrrrrrrrrrr}
\multicolumn{1}{p{2.5ex}}{\rotatebox{0}{Model}} & 
\multicolumn{1}{p{2.5ex}}{\rotatebox{\rotation}{\textbf{\textsc{Overall}}}} &
\multicolumn{1}{p{2.5ex}}{\rotatebox{\rotation}{\textsc{D-n agr}}} &
\multicolumn{1}{p{2.5ex}}{\rotatebox{\rotation}{\textsc{S-v agr}}} &
\multicolumn{1}{p{2.5ex}}{\rotatebox{\rotation}{\textsc{Ana.\ agr}}} &
\multicolumn{1}{p{2.5ex}}{\rotatebox{\rotation}{\textsc{Arg.\ str}}} &
\multicolumn{1}{p{2.5ex}}{\rotatebox{\rotation}{\textsc{Binding}}} &
\multicolumn{1}{p{2.5ex}}{\rotatebox{\rotation}{\textsc{Case}}} &
\multicolumn{1}{p{2.5ex}}{\rotatebox{\rotation}{\textsc{Ellipsis}}} &
\multicolumn{1}{p{2.5ex}}{\rotatebox{\rotation}{\textsc{Filler.\ gap}}} &
\multicolumn{1}{p{2.5ex}}{\rotatebox{\rotation}{\textsc{Irregular}}} &
\multicolumn{1}{p{2.5ex}}{\rotatebox{\rotation}{\textsc{Island}}} &
\multicolumn{1}{p{2.5ex}}{\rotatebox{\rotation}{\textsc{Local.\ atr}}} &
\multicolumn{1}{p{2.5ex}}{\rotatebox{\rotation}{\textsc{Quantifiers}}}  & 
\multicolumn{1}{p{2.5ex}}{\rotatebox{\rotation}{\textsc{NPI}}} \\  \cmidrule(lr){1-15} 
\textbf{AO-CHILDES} & & & & & & & & & & & & & & \\
\textsc{DynamicLimit-Exp} (\textuparrow) & 62.2\phantom{$^{*}$} & 50.8\phantom{$^{*}$} & 50.0\phantom{$^{*}$} & 49.6\phantom{$^{*}$} & 67.7\phantom{$^{*}$} & 58.7\phantom{$^{*}$} & 95.2\phantom{$^{*}$} & 43.1\phantom{$^{*}$} & 93.6\phantom{$^{*}$} & 52.2\phantom{$^{*}$} & 53.6\phantom{$^{*}$} & 51.3\phantom{$^{*}$} & 57.6\phantom{$^{*}$} & 85.0\phantom{$^{*}$} \\
\rowcolor{backgray}
\textsc{DynamicLimit-Exp} (\textdownarrow) & 56.5\phantom{$^{*}$} & 49.9\phantom{$^{*}$} & 49.7\phantom{$^{*}$} & 50.1\phantom{$^{*}$} & 44.7\phantom{$^{*}$} & 61.9\phantom{$^{*}$} & 70.6\phantom{$^{*}$} & 73.3\phantom{$^{*}$} & 72.0\phantom{$^{*}$} & 51.8\phantom{$^{*}$} & 61.9\phantom{$^{*}$} & 47.0\phantom{$^{*}$} & 48.1\phantom{$^{*}$} & 54.1\phantom{$^{*}$} \\
$\Delta$ (\textuparrow, \textdownarrow) & \textbf{\textcolor{blue}{5.7}}$^{*}$ & \textbf{\textcolor{blue}{0.9}}\phantom{$^{*}$} & \textbf{\textcolor{blue}{0.3}}\phantom{$^{*}$} & {-0.5}\phantom{$^{*}$} & \textbf{\textcolor{blue}{23.0}}$^{*}$ & {-3.2}$^{\dagger}$ & \textbf{\textcolor{blue}{24.6}}$^{*}$ & {-30.1}$^{\dagger}$ & \textbf{\textcolor{blue}{21.6}}$^{*}$ & \textbf{\textcolor{blue}{0.4}}\phantom{$^{*}$} & {-8.3}$^{\dagger}$ & \textbf{\textcolor{blue}{4.4}}$^{*}$ & \textbf{\textcolor{blue}{9.5}}$^{*}$ & \textbf{\textcolor{blue}{30.8}}$^{*}$ \\ \midrule
\textbf{Wikipedia} & & & & & & & & & & & & & & \\
\textsc{DynamicLimit-Exp} (\textuparrow) & 59.1\phantom{$^{*}$} & 49.8\phantom{$^{*}$} & 50.4\phantom{$^{*}$} & 46.0\phantom{$^{*}$} & 71.5\phantom{$^{*}$} & 59.3\phantom{$^{*}$} & 97.7\phantom{$^{*}$} & 37.4\phantom{$^{*}$} & 86.5\phantom{$^{*}$} & 51.1\phantom{$^{*}$} & 58.0\phantom{$^{*}$} & 60.5\phantom{$^{*}$} & 42.2\phantom{$^{*}$} & 53.9\phantom{$^{*}$} \\
\rowcolor{backgray}
\textsc{DynamicLimit-Exp} (\textdownarrow) & 52.9\phantom{$^{*}$} & 50.4\phantom{$^{*}$} & 50.1\phantom{$^{*}$} & 47.4\phantom{$^{*}$} & 68.7\phantom{$^{*}$} & 62.3\phantom{$^{*}$} & 74.4\phantom{$^{*}$} & 60.2\phantom{$^{*}$} & 44.2\phantom{$^{*}$} & 53.2\phantom{$^{*}$} & 51.7\phantom{$^{*}$} & 42.7\phantom{$^{*}$} & 40.6\phantom{$^{*}$} & 42.2\phantom{$^{*}$} \\
$\Delta$ (\textuparrow, \textdownarrow) & \textbf{\textcolor{blue}{6.1}}$^{*}$ & {-0.6}\phantom{$^{*}$} & \textbf{\textcolor{blue}{0.3}}\phantom{$^{*}$} & {-1.4}\phantom{$^{*}$} & \textbf{\textcolor{blue}{2.9}}\phantom{$^{*}$} & {-3.0}\phantom{$^{*}$} & \textbf{\textcolor{blue}{23.3}}$^{*}$ & {-22.8}$^{\dagger}$ & \textbf{\textcolor{blue}{42.3}}$^{*}$ & {-2.2}\phantom{$^{*}$} & \textbf{\textcolor{blue}{6.3}}$^{*}$ & \textbf{\textcolor{blue}{17.8}}$^{*}$ & \textbf{\textcolor{blue}{1.7}}\phantom{$^{*}$} & \textbf{\textcolor{blue}{11.7}}$^{*}$ \\
\bottomrule
\end{tabular}
\end{adjustbox}
\caption{Performance difference when changing the direction of the cognitive constraints in \textsc{DynamicLimit-Exp}. $^{*}$ and $^{\dagger}$ indicate items where \textsc{DynamicLimit-Exp} (\textdownarrow) performed significantly better or worse than \textsc{DynamicLimit-Exp} (\textuparrow), respectively (z-test for proportions, p~<~0.05).}
\label{tab:vs_reverse}
\end{table*}

\subsection{Baselines}
We prepared the following three baseline models to precisely analyze the learning effects of different working memory limitation strategies:

\begin{itemize}
    \item \textbf{\textsc{NoLimit}}: A model with no memory constraints. Working memory remains constant from the early stages of training, simulating the mature working memory observed post-adolescence. This configuration is equivalent to a vanilla GPT-2~\cite{radford2019language}.
    \item \textbf{\textsc{StaticLimit}}: A model applying standard ALiBi~\cite{press2022train} during attention score calculation, where memory constraints remain fixed throughout training.
    \item \textbf{\textsc{DynamicLimit-Linear}}: A model in which the ALiBi slope \(m\) decreases linearly over the course of training.
\end{itemize}

To ensure a fair comparison between the linear and exponential growth curves of working memory, we controlled the initial and final values of working memory capacity \(w_t\) in \textsc{DynamicLimit-Linear} and \textsc{DynamicLimit-Exp} to be as similar as possible.  
Specifically, we set the number of training epochs to 10 and configured both models with an initial slope of \(m = 1.0\) and a final slope of \(m = 0.0\).  
Figure~\ref{fig:wm_curves} illustrates the trajectory of working memory capacity for each model.  
All models were trained using three different seeds, and we report the average results across these runs.

\subsection{Results}
\label{subsec:main_results}
\paragraph{Developmentally-plausible working memory shapes the CP for \lone acquisition}
Table~\ref{tab:main_result} presents the accuracy of each model trained on the AO-CHILDES.  
Compared to \textsc{NoLimit} and \textsc{StaticLimit}, which do not account for developmental changes in working memory, \textsc{DynamicLimit-Linear} and \textsc{DynamicLimit-Exp}, which simulate its gradual growth, achieve significantly higher overall performance.  
Among them, \textsc{DynamicLimit-Exp} attains the highest overall accuracy, supporting the effectiveness of a cognitively plausible mechanism.  
The comparable performance of \textsc{StaticLimit} to \textsc{NoLimit} suggests that the gradual introduction of working memory constraints throughout training is crucial, rather than their static application.  
These results indicate that \textsc{DynamicLimit-Exp} effectively replicates the CP effect observed in human \lone acquisition.

\paragraph{The CP depends on the child's learning algorithm, not the input stimulus}
Table~\ref{tab:wiki_result} presents the accuracy of models trained on Wikipedia, showing trends similar to those observed in Table~\ref{tab:main_result}, where the models were trained on AO-CHILDES. 
Specifically, \textsc{DynamicLimit-Linear} and \textsc{DynamicLimit-Exp} outperform \textsc{NoLimit} and \textsc{StaticLimit} in overall accuracy, with \textsc{DynamicLimit-Exp} achieving the highest performance, further supporting the efficacy of incorporating developmental working memory constraints.
These findings suggest that the CP effect does not depend solely on exposure to specific linguistic stimuli (e.g., CDS) but rather on the learning algorithm itself, which mirrors human cognitive development. 
This result aligns with existing research~\cite{feng-etal-2024-child,padovani2025childdirectedlanguagedoesconsistently}, which have reported that CDS is not uniquely valuable for training LMs.
This finding suggests that our method is applicable to LLM pretraining, as they typically use non-CDS datasets such as Common Crawl and Wikipedia~\cite{touvron2023llamaopenefficientfoundation}.

\begin{figure*}[t]
 \centering
  \includegraphics[width=0.9\linewidth]{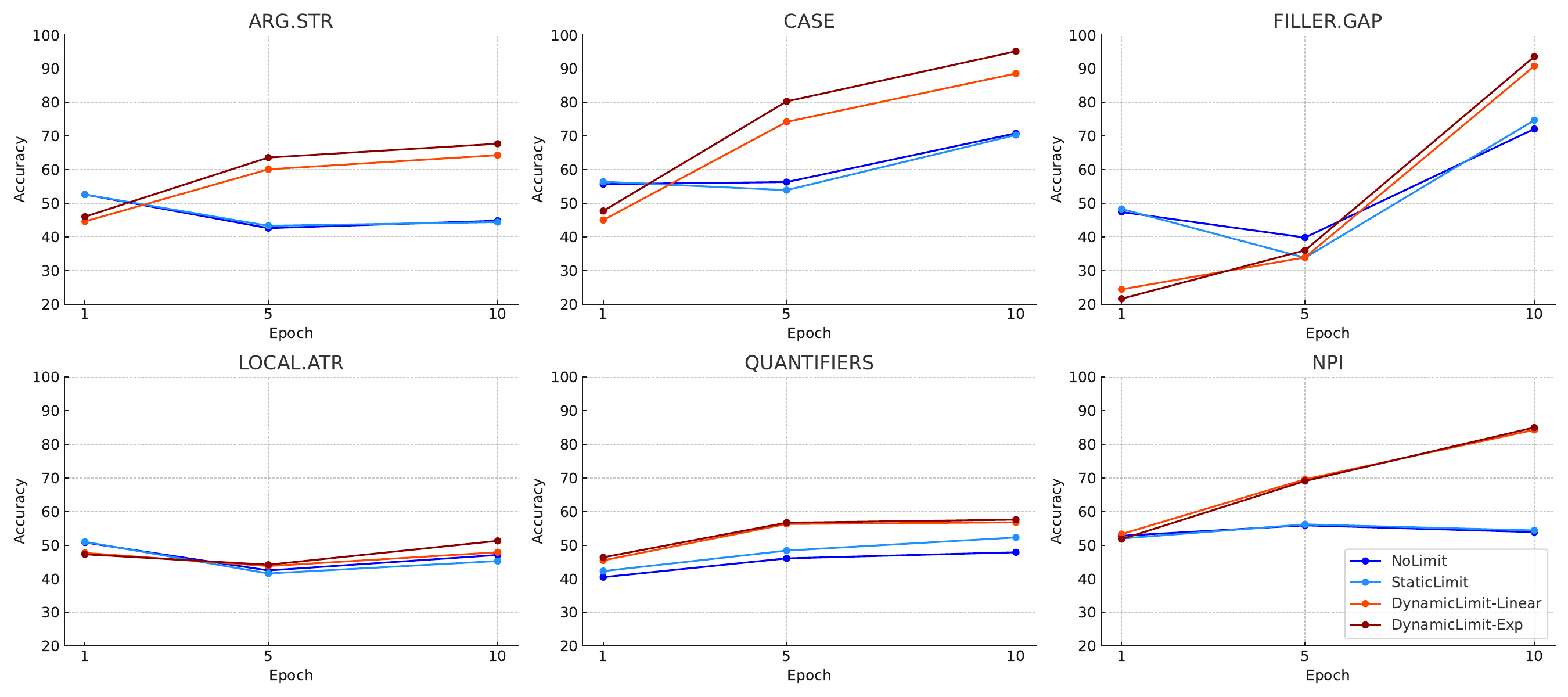}
 \caption{Accuracy trajectories over training epochs (1, 5, 10) for six grammatical categories that showed significant final-stage improvements with developmental constraints.}
 \label{fig:evolution}
\end{figure*}

\section{Analysis}

\subsection{Testing the ``Less-is-more'' Hypothesis with Reversed Cognitive Constraints}
\label{subsec:test_reversed}
A key question arising from the results (\S\ref{sec:experiment}) is whether \textsc{DynamicLimit-Exp}'s superior performance stems from the ``Less-is-more'' hypothesis~\cite{NEWPORT199011}—i.e., the gradual growth of working memory—or from unintended side effects. 
In other words, does the gradual \textit{change} in working memory enhance information capacity, dynamically shifting the model's focus across epochs and ultimately aiding rule generalization?
To test this, we introduce a cognitively \textit{implausible} language model, referred to as~``\textsc{DynamicLimit-Exp} (\textdownarrow)'', which shares the same slope trajectory as our proposed~\textsc{DynamicLimit-Exp} (\textuparrow)~\footnote{This section adopts this notation for simplicity.} but with its direction reversed, such that working memory capacity decreases over time.
Specifically, DynamicLimit-Exp (\textuparrow) is set to \( m_0 = 1.0, r = 0.6 \) (the same setting as in \S\ref{sec:experiment}), while DynamicLimit-Exp~(\textdownarrow) is set to \( m_0 = 0.01, r = 1.668 \) to achieve a nearly symmetrical curve.\footnote{Since setting the initial slope \( m_0 = 0.0 \) prevents \( w_t \) from being updated in Equation (2), we set it this way for computational reasons.
}

Table~\ref{tab:vs_reverse} provides evidence supporting the Less-is-more hypothesis, as~\textsc{DynamicLimit-Exp} (\textuparrow) consistently outperformed the cognitively implausible~\textsc{DynamicLimit-Exp} (\textdownarrow).  
The observed performance gap, particularly in grammatical items requiring both local and non-local dependencies (e.g., CASE, ARG. STR, and FILLER-GAP), suggests that the gradual growth of working memory is crucial for grammatical learning and generalization, as it enables the early extraction of basic patterns followed by the progressive acquisition of complex rules.  
These findings indicate that the superior performance of~\textsc{DynamicLimit-Exp} (\textuparrow) is primarily driven by the developmental trajectory of working memory growth rather than unintended side effects of dynamic shifts in memory focus.

Incidentally, from the series of experimental results, along with those in \S\ref{sec:experiment} (Table~\ref{tab:main_result} and \ref{tab:wiki_result}), \textsc{NoLimit} and \textsc{DynamicLimit-Exp} (\textdownarrow) consistently outperform \textsc{DynamicLimit-Exp} (\textuparrow) in~\textsc{ELLIPSIS}, as exemplified by the following cases:
\eenumsentence[2]{ \item \ac Mark fixed one \textbf{worn} canal, and Roger \nextline{fixed more.} \item \un Mark fixed one canal, and Roger fixed \nextline{more \textbf{worn}.} }
Since resolving \textsc{ELLIPSIS} involves maintaining long-range dependencies, \textsc{DynamicLimit-Exp} (\textuparrow) may struggle due to its initial memory constraints. 
This suggests that grammatical items like \textsc{ELLIPSIS} require substantial memory from the early stages of training, and thus, our proposed method may not be optimal for learning such structures. 
Alternative workarounds, such as dynamically adjusting memory allocation or hybrid approaches, may be necessary to address this limitation.

\subsection{Tracking Developmental Gains Across Training}
To more directly support our claim that developmentally guided learning simulates a CP, we examine how model performance unfolds over time—not just at the endpoint but at intermediate stages as well. This addresses the need for stage-by-stage comparisons raised by prior evaluations.

Figure~\ref{fig:evolution} tracks accuracy at Epochs 1, 5, and 10 for six grammatical categories selected based on statistically significant improvements observed in Table~\ref{tab:main_result}.
At the early stage (Epoch 1), models with larger or fixed memory (\textsc{NoLimit}, \textsc{StaticLimit}) perform better. However, the developmentally constrained model, \textsc{DynamicLimit-Exp}, shows steady gains over time, ultimately surpassing these baselines by Epoch 10 in multiple categories. The improvement is especially pronounced in \textsc{CASE} and \textsc{FILLER.GAP}, highlighting a pattern of late-stage acceleration.
These results suggest that incrementally increasing memory capacity over training acts as a beneficial inductive bias, enabling the model to generalize more effectively from limited early experience—consistent with the hypothesized role of a CP in human language acquisition.

\begin{figure*}[t]
    \begin{tabular}{c}
    
  \begin{minipage}[]{0.5\linewidth}
    \centering
    \includegraphics[width=0.9\linewidth]{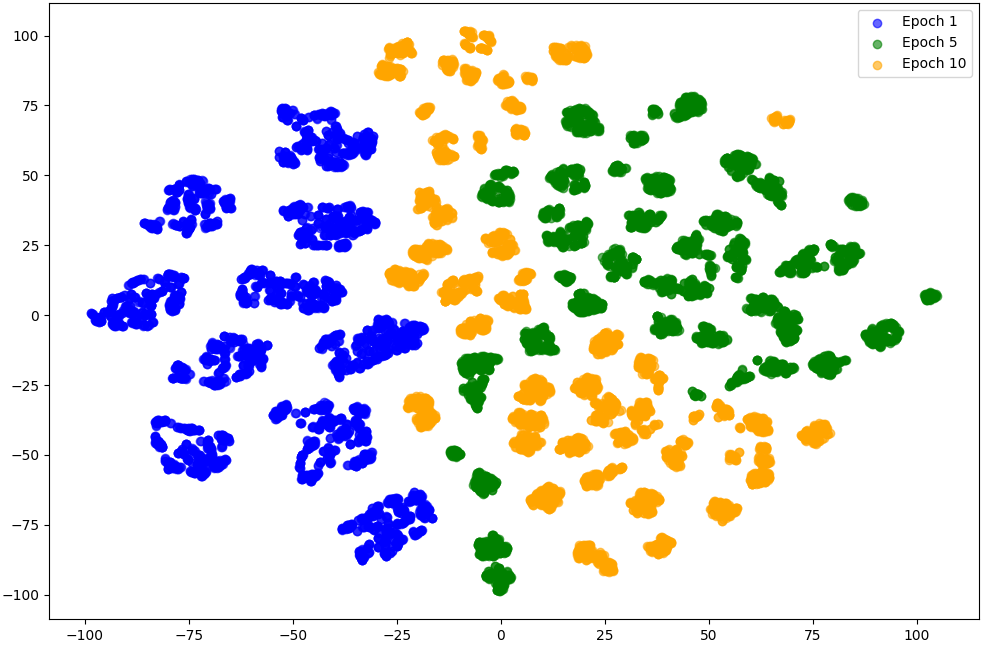}    
    \subcaption{\textsc{NoLimit}}\label{fig:vanilla}
  \end{minipage}

    \begin{minipage}[]{0.5\linewidth}
    \centering
    \includegraphics[width=0.9\linewidth]{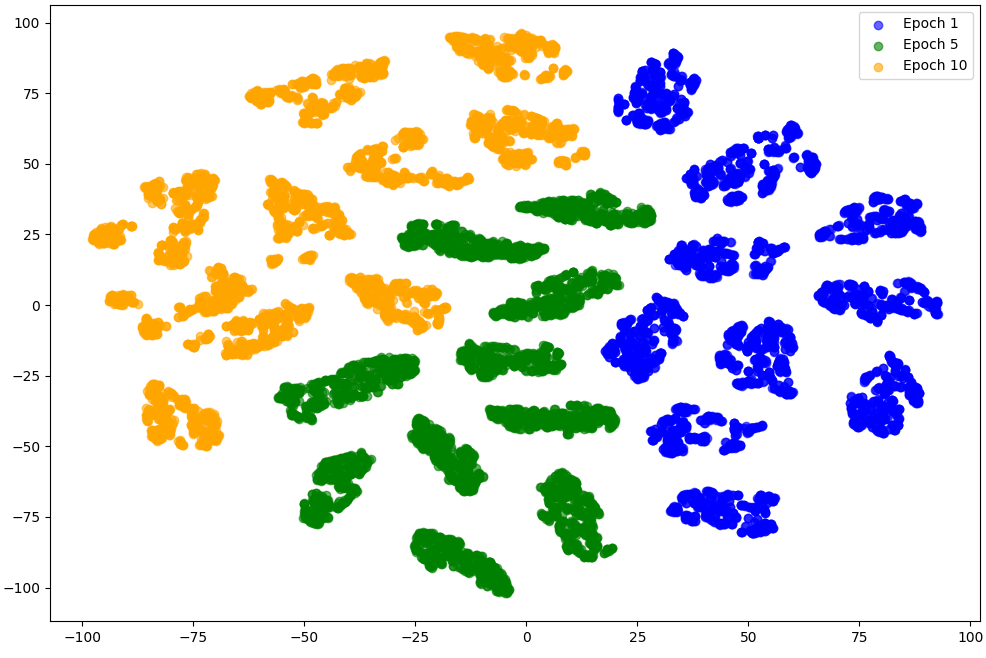}    
    \subcaption{\textsc{DynamicLimit-Exp}}\label{fig:exp}
  \end{minipage}

  \end{tabular}
  \caption{Embedded space at each learning stage for \textsc{NoLimit} and \textsc{DynamicLimit-Exp} (\textsc{Filler.\ gap})}
  \label{fig:embedded_space}
\end{figure*}

\begin{table}[t!]
\centering
\small
\tabcolsep=2.5pt 
\begin{tabular}{lcccccc}
\toprule
       & \multicolumn{3}{c}{Entropy}              & \multicolumn{3}{c}{Mean Distance}        \\ \cmidrule(r){2-4}\cmidrule(r){5-7}
Epoch  & 1     & 5     & 10    & 1-5   & 5-10  & 1-10  \\ \midrule
NoLimit           & 5.36  & 5.17  & 5.19  & 91.30 & 28.50  & 66.28 \\
DynamicLimit-Exp  & 5.40  & 5.30  & 5.39  & 69.25 & 70.63  & 101.92 \\ \bottomrule
\end{tabular}
\caption{Embedded space analysis of \textsc{NoLimit} and \textsc{DynamicLimit-Exp} at each stage: distribution diversity and distribution distance.}
\label{tab:cluster_analysis}
\end{table}

\subsection{Learning to Represent: The Cognitive Effect of Memory Expansion}

We analyze representational change by tracking embedding diversity within epochs and shifts between epochs.
Specifically, we consider two complementary aspects of representational change: 
(i) the diversity or dispersion of embeddings \textit{within} each epoch, which reflects the isotropy and expressiveness of the representation space at a given time; and 
(ii) the amount of shift in embeddings \textit{between} epochs, which captures the degree of representational update and learning progress over time.

Figure~\ref{fig:embedded_space} visualizes t-SNE projected embeddings for the \textsc{FILLER.GAP} category, where \textsc{DynamicLimit-Exp} showed clear gains over baselines (\S\ref{subsec:main_results}, \S\ref{subsec:test_reversed}).
In the \textsc{NoLimit} model (Figure~\ref{fig:embedded_space}a), embedding clusters expand from Epoch 1 to 5 but subsequently contract and overlap by Epoch 10.
This pattern reflects a reduction in within-epoch diversity (i.e., lower isotropy) and minimal between-epoch shift, indicating that the representations stagnate and fail to evolve structurally as training progresses.
In contrast, \textsc{DynamicLimit-Exp} (Figure~\ref{fig:embedded_space}b) produces more structured trajectories: clusters remain well-separated within epochs and continue to shift meaningfully across epochs.
This suggests not only sustained representational plasticity but also a finer-grained encoding of syntactic distinctions over time.

To quantify these trends, Table~\ref{tab:cluster_analysis} reports entropy (capturing embedding dispersion) and mean Euclidean distance between clusters (capturing separation).\footnote{The Appendix~\ref{appendix:statistical_measures} shows how to calculate each measure.}
The \textsc{NoLimit} model shows a drop in entropy and a plateau in inter-cluster distance after Epoch 5, consistent with representational collapse.
Meanwhile, \textsc{DynamicLimit-Exp} maintains higher entropy and exhibits a steady increase in distance, consistent with ongoing structural refinement.
These findings indicate that developmentally guided memory expansion helps preserve expressive, isotropic, and well-separated embedding spaces—properties that support better generalization in language models~\cite{diehl-martinez-etal-2024-mitigating}.\footnote{Similar trends were found for \textsc{CASE}; see Appendix~\ref{appendix:cluster_analysis}.}

\subsection{Influence of Input Stimulus Length}

\begin{table}[t!]
\small
\centering
\begin{tabular}{lrr} \toprule
Dataset & \textsc{NoLimit} & \textsc{DynamicLimit-Exp} \\  \midrule
{[}5,10{]}   & \textbf{47.2}\phantom{$^{*}$}     & 46.8\phantom{$^{*}$}  \\
{[}11,50{]}  & 47.0\phantom{$^{*}$}              & \textbf{58.7$^{*}$}   \\
{[}51,100{]} & 40.6\phantom{$^{*}$}              & \textbf{42.5$^{*}$}   \\
{[}101, 150{]} & 37.3\phantom{$^{*}$}            & \textbf{40.8$^{*}$}   \\ \bottomrule    
\end{tabular}
\caption{Accuracy in Zorro when the length of the sentence is changed. $^{*}$ indicates a statistically significant differences (p~<~0.05).
}
\label{tab:length}
\end{table}

We analyze how sentence length affects the performance of \textsc{NoLimit} and \textsc{DynamicLimit-Exp}. 
To assess their adaptability, we created four Wikipedia-based datasets, each with 500,000 sentences in length ranges: [5,10], [11,50], [51,100], and [101,150].

The results in Table~\ref{tab:length} reveal notable differences in model performance. 
For shorter sentences in the [5,10] range, \textsc{NoLimit} achieves slightly higher accuracy compared to \textsc{DynamicLimit-Exp}. 
However, in the [11,50] range, \textsc{DynamicLimit-Exp} significantly outperforms \textsc{NoLimit}, achieving 58.7 compared to 47.0. 
This suggests that \textsc{DynamicLimit-Exp} excels at handling moderately long sentences, likely due to its ability to dynamically adjust working memory.
For longer sentences in the [51,100] and [101,150] ranges, \textsc{DynamicLimit-Exp} consistently outperforms \textsc{NoLimit}.
These findings highlight the benefits of dynamic working memory expansion in facilitating rule generalization and contextual adaptation across diverse sentence lengths. 
While \textsc{NoLimit} exhibits competitive performance on short sentences, its stagnation on longer sentences underscores its limited ability to generalize complex patterns. 
Conversely, \textsc{DynamicLimit-Exp}'s consistent performance across varying sentence lengths supports its suitability for grammatical items requiring the processing of both short and long contexts.

\section{Conclusion}
This study proposed a method for integrating the developmental trajectory of human working memory into the training process of LMs, inspired by the \textit{Less-is-More} hypothesis. 
The proposed method, \textsc{DynamicLimit-Exp}, initially restricts working memory and gradually relaxes it exponentially during training.
Experiments on both AO-CHILDES and Wikipedia showed that \textsc{DynamicLimit-Exp} improves grammatical learning efficiency compared to conventional methods without memory constraints or with static memory constraints.
These findings suggest a promising direction for building more data-efficient LMs by leveraging cognitively inspired inductive biases.

Beyond its engineering implications, this study also offers insight into the cognitive mechanisms underlying \lone acquisition. 
While our results do not directly demonstrate that working memory development is necessary for human learners, they serve as a computational-level plausibility test~\cite{Marr1982}, showing that the hypothesized link between cognitive constraints and rule learning, central to the \textit{Less-is-More} hypothesis, can be instantiated in artificial learners.
Combining the observed learning efficiency gains and the cognitive plausibility of our approach, we support the hypothesis-generating idea that such developmental constraints may plausibly aid human language acquisition as well.

\section*{Acknowledgments}
We thank the anonymous reviewers for their helpful comments and suggestions.
We are also grateful to Akiyo Fukatsu for her insightful comments and the Computational Models of Language (CoLa) Reading Group at UC Irvine for their constructive feedback.
This work was supported by JSPS KAKENHI Grant Number 24H00087, JST PRESTO Grant Number JPMJPR21C2, Grant-in-Aid for JSPS Fellows JP24KJ0800, JST SPRING Grant Number JPMJSP2108.

\section*{Limitations}
\paragraph{Scalability.}
One limitation of this study is the constrained scale of the experimental setup.  
The primary goal of this study is to computationally replicate the CP in \lone acquisition, as discussed in cognitive science~\cite{lenneberg1967biological,FROMKIN197481,curtiss1977genie,JOHNSON198960}.  
Following previous studies~\cite{huebner-etal-2021-babyberta,haga-etal-2024-modeling}, we designed the experiment to be as ecologically valid as possible by training an SLM using CDS.  
While this controlled setting allows for a more precise analysis and simulation of the Less-is-More hypothesis, it remains unclear how our findings contribute to the data efficiency of LLMs.  
The experimental results with Wikipedia (Table~\ref{tab:wiki_result}, \ref{tab:vs_reverse}, \ref{tab:length}) provide a promising outlook in this direction, but further investigation with larger models and datasets is necessary to determine the effectiveness and limitations of the proposed approach.  

\paragraph{Language.}
In this experiment, we investigated the replication of the CP effect in \lone acquisition using English.  
However, since the CP effect is observed across various languages~\cite{Patkowski1980,JOHNSON198960}, it remains to be tested whether the proposed approach is effective in multilingual environments.  
To our knowledge, there is currently no targeted syntactic evaluation specifically designed for CDS across different languages, such as Zorro.  
Zorro was developed based on BLiMP~\cite{warstadt-etal-2020-blimp-benchmark}, an adult-oriented targeted syntactic evaluation for English, and recent studies have proposed multilingual versions of BLiMP (e.g., JBLiMP~\cite{someya-oseki-2023-jblimp} for Japanese and CLiMP~\cite{xiang-etal-2021-climp} for Chinese).  
Therefore, developing CDS-specific versions based on these multilingual BLiMPs could help address this limitation.

\bibliography{custom}

\clearpage
\appendix

\section{Details of the Training Configuration for the Base Models}
\label{appendix:detailed_settings}
Table~\ref{table:hyperparameters} shows the training settings of the base model.
For the experiment, a single NVIDIA RTX A5000 (24GB) GPU was used, and the training time for each run was approximately one hour.

\begin{table}[h!]
\centering
\small
\tabcolsep=2pt 
\begin{tabular}{|l|l|}
\hline
\textbf{Hyperparameter}       & \textbf{Value}                  \\ \hline
Model Architecture            & GPT-2     \\ \hline
Number of Layers              & 4                               \\ \hline
Number of Attention Heads     & 4                               \\ \hline
Embedding Dimension           & 256                             \\ \hline
Dropout Rate                  & 0.1                             \\ \hline
Learning Rate (\(\eta\))      & \(5 \times 10^{-6}\)            \\ \hline
Weight Decay                  & 0.01                            \\ \hline
Batch Size                    & 512                             \\ \hline
Gradient Accumulation Steps   & 2                               \\ \hline
Total Epochs                  & 20                              \\ \hline
Maximum Sequence Length       & 32                          \\ \hline
Learning Rate Scheduler       & Cosine with Restarts         \\ \hline
Warm-up Steps                 & 10\% of Total Steps             \\ \hline
Optimizer                     & AdamW                            \\ \hline
Optimizer Parameters          & $\beta = (0.9, 0.999)$, $\epsilon = 1e−08$ \\ \hline
Tokenizer                     & Trained on CHILDES \\ \hline
Early Stopping Tolerance      & 1 Epoch                         \\ \hline
Evaluation Metric             & Perplexity                      \\ \hline
\end{tabular}
\caption{Training Configuration (Hyperparameters) for the GPT-2 Model.}
\label{table:hyperparameters}
\end{table}

\section{Distribution of the Datasets}
\label{appendix:dist}

\begin{figure*}[h!]
    \begin{tabular}{c}

  \begin{minipage}[]{0.5\linewidth}
    \centering
    \includegraphics[width=1.0\linewidth]{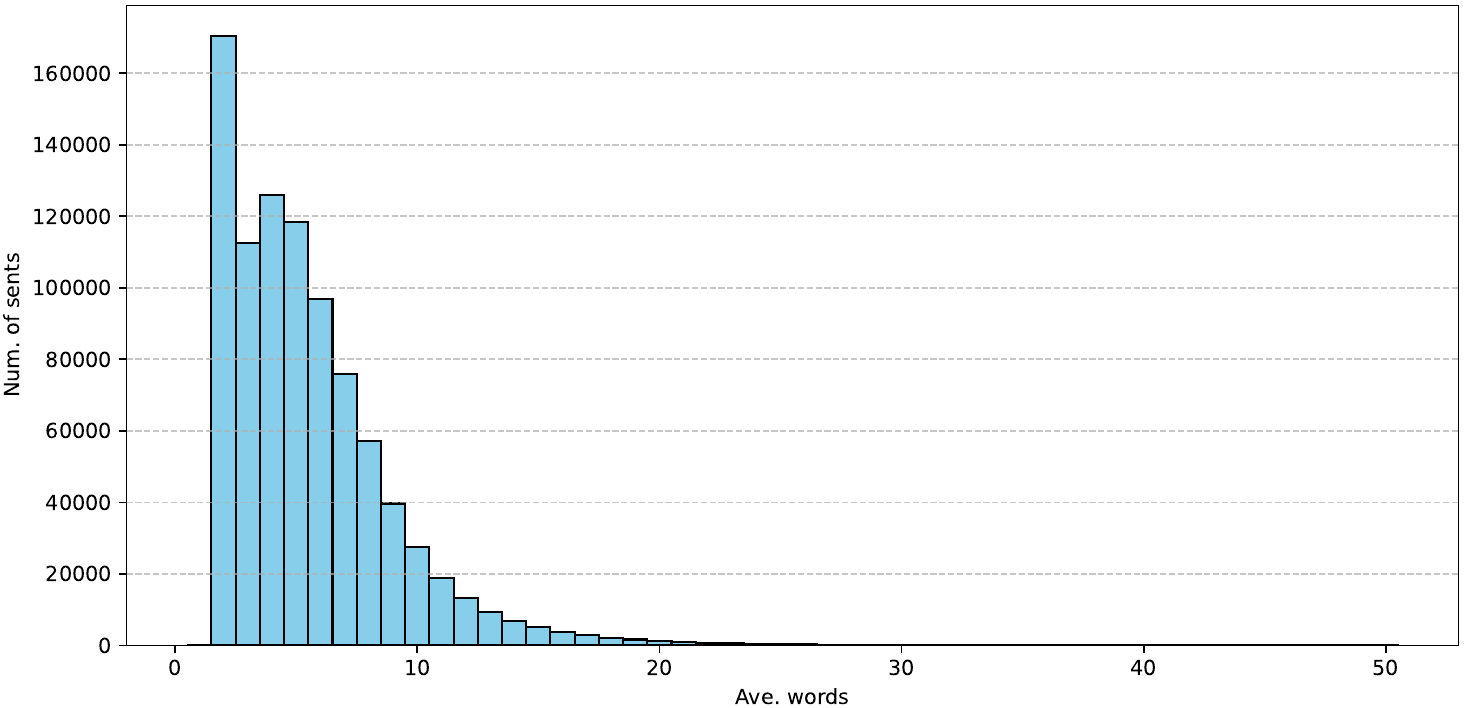}    
    \subcaption{AO-CHILDES}\label{fig:dist_aochildes}
  \end{minipage}

    \begin{minipage}[]{0.5\linewidth}
    \centering
    \includegraphics[width=1.0\linewidth]{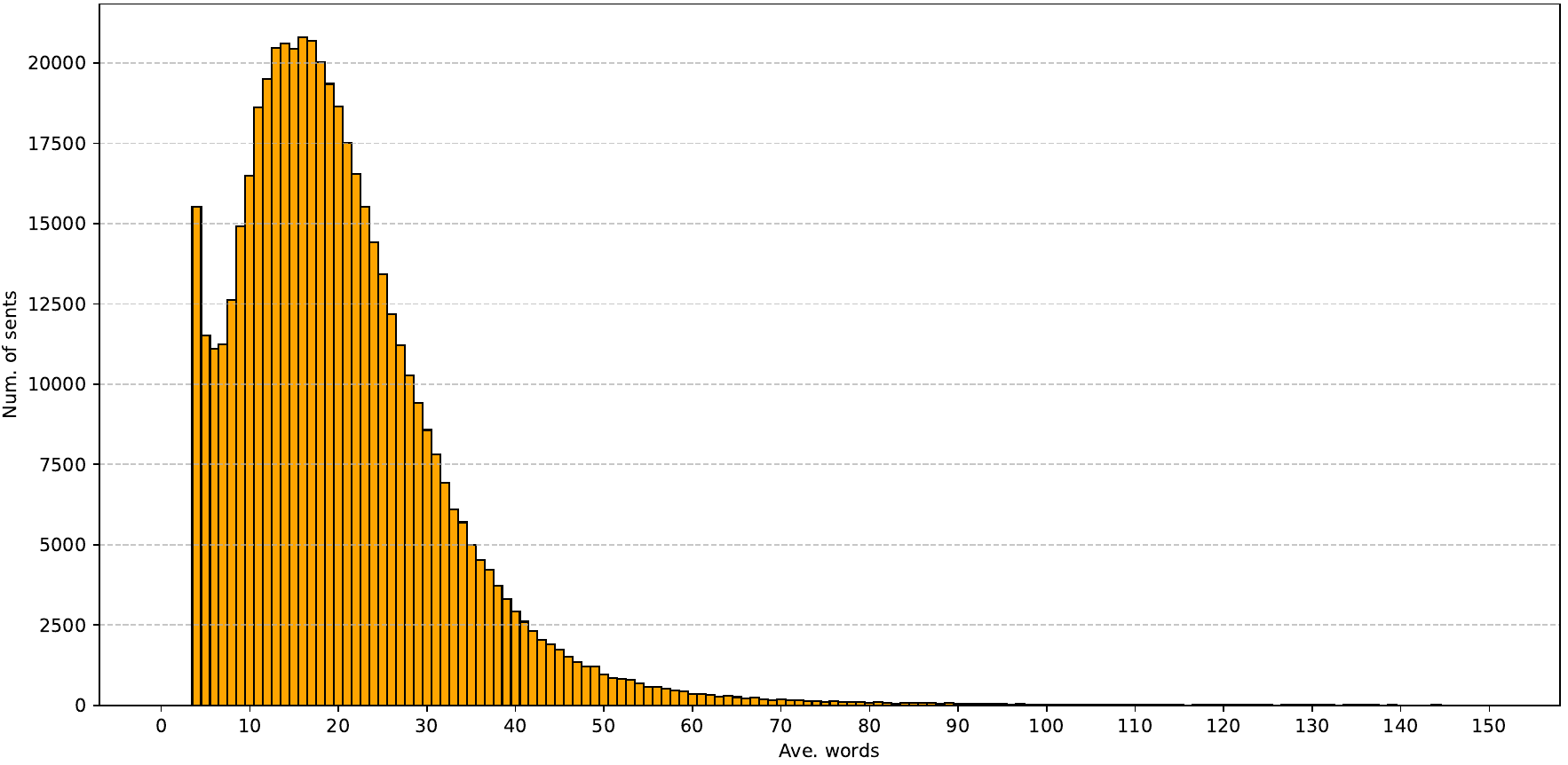}    
    \subcaption{Wikipedia}\label{fig:dist_wikipedia}
  \end{minipage}

  \end{tabular}
  \caption{Word distribution of the AO-CHILDES and Wikipedia datasets used in the experiment}
  \label{fig:dist_dataset}
\end{figure*}

Figure~\ref{fig:dist_dataset} shows the sentence length distribution for the AO-CHILDES and Wikipedia datasets used in this experiment. As can be seen from the figure, AO-CHILDES, by its nature, contains more short sentences than Wikipedia.

\begin{table}[t!]
\centering
\small
\tabcolsep=2.5pt 
\begin{tabular}{lcccccc}
\toprule
       & \multicolumn{3}{c}{Entropy}              & \multicolumn{3}{c}{Mean Distance}        \\ \cmidrule(r){2-4}\cmidrule(r){5-7}
Epoch  & 1     & 5     & 10    & 1-5   & 5-10  & 1-10  \\ \midrule
NoLimit           & 5.30  & 5.23  & 5.30  & 75.47 & 12.26  & 87.62 \\
DynamicLimit-Exp  & 5.29  & 5.30  & 5.34  & 59.91 & 37.68  & 97.59 \\ \bottomrule
\end{tabular}
\caption{Embedded space analysis of \textsc{NoLimit} and \textsc{DynamicLimit-Exp} at each stage: cluster expansion, distribution diversity, and distribution distance.}
\label{tab:cluster_analysis_case}
\end{table}

\section{Details of Grammatical Items in Zorro}
\label{appendix:zorro_items}

\begin{table*}[t!]
\scriptsize
\centering
\tabcolsep=2pt 
\scalebox{0.85}{ 
\begin{tabular}{p{2.2cm} p{3.5cm} p{6cm} p{6cm}}\toprule
Category &
  Subcategory &
  Acceptable Sentence &
  Unacceptable Sentence \\ \midrule
 &
  noun-across\_1\_adjective &
  \textit{look at this purple \textbf{thing} .} &
  \textit{look at this purple \textbf{things} .} \\ 
\multirow{-2}{*}{D-N AGR} &
  noun-between\_neighbors &
  \textit{this \textbf{color} must be white .} &
  {\textit{this \textbf{colors} must be white .}} \\ \midrule
{ } & {verb-across\_prepositional\_phrase} & 
\textit{the \textbf{lie} on the foot is flat .} & 
\textit{the \textbf{lies} on the foot is flat .} \\
{ } &
  verb-across\_relative\_clause &
  \textit{the \textbf{book} that i like is poor .} &
  \textit{the \textbf{books} that i like is poor .} \\
{ } &
  verb-in\_question\_with\_aux &
  \textit{where does the \textbf{horse} go ?} &
  \textit{where does the \textbf{horses} go ?} \\ 
\multirow{-4}{*}{{S-V AGR}} &
  verb-in\_simple\_question &
  \textit{where is the \textbf{way} ?} &
  \textit{where is the \textbf{ways} ?} \\  \midrule
ANA.AGR &
  pronoun\_gender &
  \textit{will Mark want \textbf{himself} ?} &
  \textit{will Mark want \textbf{herself} ?} \\ \midrule
 &
  dropped\_argument &
  \textit{\textbf{give me the poor boat} .} &
  \textit{\textbf{the poor boat gives me} .} \\
 &
  swapped\_arguments &
  \textit{\textbf{he made the slave her label} .} &
  \textit{\textbf{the slave made her label he} .} \\   
\multirow{-3}{*}{ARG.STR} &
  transitive &
  \textit{Philip \textbf{thinks} .} &
  \textit{Philip \textbf{affected} .} \\ \midrule
BINDING &
  principle\_a &
  \textit{Ben thinks about himself \textbf{calling} this fuel .} &
  \textit{Ben thinks about himself \textbf{called} this fuel .} \\ \midrule
CASE &
  subjective\_pronoun &
  \textit{\textbf{i brought the wolf} my hill .} &
  \textit{\textbf{the wolf brought i} my hill .} \\ \midrule
ELLIPSIS &
  n\_bar &
  \textit{Mark fixed one \textbf{worn} canal and Roger fixed more .} &
  \textit{Mark fixed one canal and Roger fixed more \textbf{worn} .} \\ \midrule
 &
  wh\_question\_object &
  \textit{Laura married the dinner \textbf{that the wolf could close} .} &
  \textit{Laura married \textbf{what} the dinner \textbf{could close the wolf} .} \\
\multirow{-2}{*}{FILLER.GAP} &
  wh\_question\_subject &
  \textit{Laura ended the finger \textbf{that} can make boats .} &
  \textit{Laura ended \textbf{who} the finger can make boats .} \\ \midrule
IRREGULAR &
  verb &
  \textit{Michael \textbf{chose} the good one some time ago .} &
  \textit{Michael \textbf{chosen} the good one some time ago .} \\ \midrule
 &
  adjunct\_island &
  \textit{who should William have \textbf{without watching the baby} ?} &
  \textit{who should William have \textbf{the baby without watching} ?} \\
\multirow{-2}{*}{ISLAND} &
  coordinate\_structure\_constraint &
  \textit{who must Philip \textbf{and the dinosaur turn} ?} &
  \textit{who must Philip \textbf{turn and the dinosaur} ?} \\ \midrule
LOCAL.ATR & 
  in\_question\_with\_aux &
  \textit{is the whale \textbf{getting} the person ?} &
  \textit{is the whale \textbf{gets} the person ?} \\ \midrule
 &
  matrix\_question &
  \textit{\textbf{\textbf{does her boat} ever play with the growth ?}} &
  \textit{\textbf{\textbf{her boat} does ever play with the growth ?}} \\
\multirow{-2}{*}{NPI} &
  only\_npi\_licensor &
  \textit{\textbf{\textbf{only} Mark ever finds some suit .}} &
  \textit{\textbf{\textbf{even} Mark ever finds some suit .}} \\ \midrule
 &
  existential\_there &
  \textit{there are \textbf{many} books about soft birds .} &
  \textit{there are \textbf{most} books about soft birds .} \\
\multirow{-2}{*}{QUANTIFIERS} &
  superlative &
  \textit{no pig could stand on top of \textbf{more than} six days .} &
  \textit{no pig could stand on top of \textbf{at least} six days .} \\ \bottomrule
\end{tabular}
}
\caption{Explanation of each grammatical category in Zorro.}\label{tab:zorro_minimal-pair}
\end{table*}

Table~\ref{tab:zorro_minimal-pair} shows the full list of grammatical categories in Zorro. 
Examples are taken from Table 5 in the original paper~\cite{huebner-etal-2021-babyberta}.

\section{Analysis of Distributional Changes in t-SNE Space Across Training Epochs}
\label{appendix:statistical_measures}
This section explains in detail the analysis of the entropy and average distance of embeddings projected into the t-SNE space for different learning epochs.

\subsection{Entropy Calculation}
To quantify the distribution of embeddings, a 2D histogram is constructed using a fixed grid (\( 50 \times 50 \) bins). 
The probability distribution \( P \) is obtained by normalizing the histogram. The entropy is then computed as:

\begin{equation}
H(P) = - \sum_{i} P_i \log P_i,
\end{equation}
where \( P_i \) is the probability of each bin. 
Higher entropy suggests a more uniform distribution, whereas lower entropy indicates clustering.

\subsection{Mean Distance Between Epochs}
To analyze shifts in embedding distributions across epochs, we compute the Euclidean distance between the mean embedding vectors of different epochs:

\begin{equation}
D(X, Y) = \| \mu_X - \mu_Y \|,
\end{equation}
where \( \mu_X \) and \( \mu_Y \) are the mean vectors at different epochs. 
Larger distances imply greater shifts in the learned representation.

\section{Development of Feature Extraction Capabilities in \textsc{CASE}}
\label{appendix:cluster_analysis}

\begin{figure*}[h!]
    \begin{tabular}{c}

  \begin{minipage}[]{0.5\linewidth}
    \centering
    \includegraphics[width=0.9\linewidth]{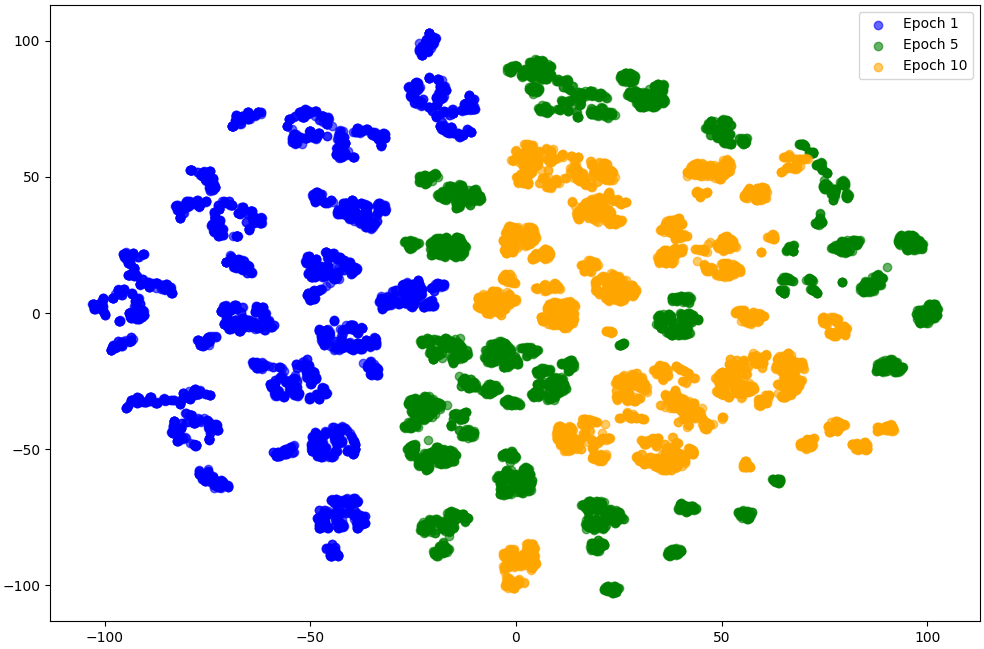}    
    \subcaption{\textsc{NoLimit}}\label{fig:vanilla}
  \end{minipage}

    \begin{minipage}[]{0.5\linewidth}
    \centering
    \includegraphics[width=0.9\linewidth]{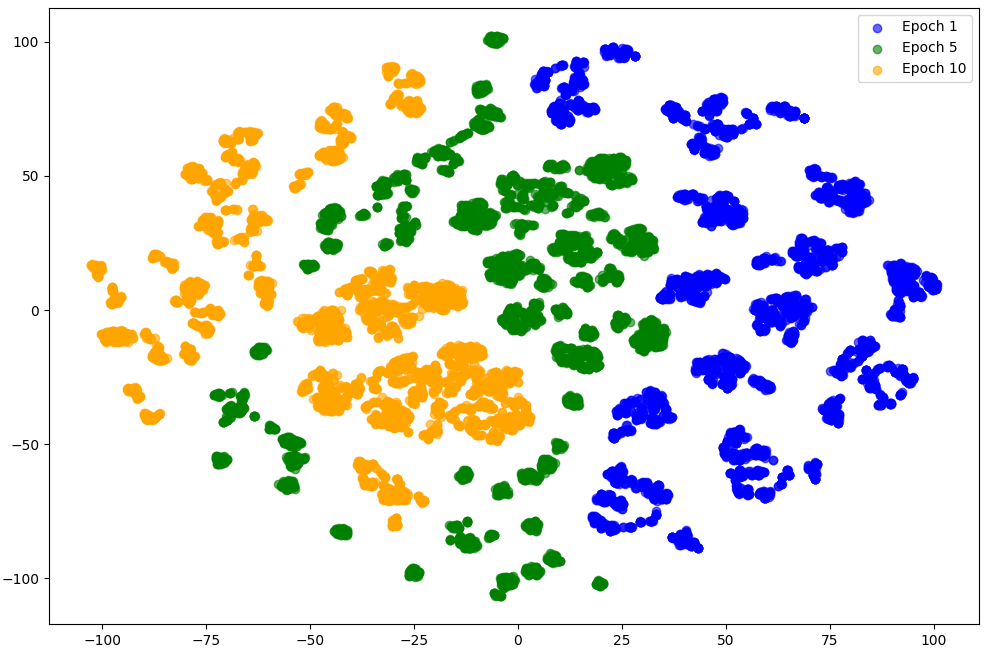}    
    \subcaption{\textsc{DynamicLimit-Exp}}\label{fig:exp}
  \end{minipage}

  \end{tabular}
  \caption{Embedded space at each learning stage for \textsc{NoLimit} and \textsc{DynamicLimit-Exp} (\textsc{CASE})}
  \label{fig:embedded_space_case}
\end{figure*}

Figure~\ref{fig:embedded_space_case} visualizes the clustering structure of final layer embeddings using t-SNE for \textsc{CASE}.
The embedding space visualizations reveal distinct patterns between~\textsc{NoLimit} and~\textsc{DynamicLimit-Exp} across training epochs. 
In~\textsc{NoLimit}, the embedding clusters expand between Epoch 1 and Epoch 5 but contract significantly by Epoch 10, suggesting stagnation in representation learning. 
In contrast,~\textsc{DynamicLimit-Exp} maintains structured evolution throughout training, with well-separated clusters that reflect progressive refinement.

Table~\ref{tab:cluster_analysis_case} shows the embedded space analysis.
Regarding \textbf{entropy}, \textsc{NoLimit} shows a slight decrease over time (Epoch 1 $\xrightarrow{}$ Epoch 5), reflecting reduced distribution diversity as training progresses. 
In contrast, \textsc{DynamicLimit-Exp} maintains or slightly increases entropy, suggesting a balanced emphasis on both basic patterns and diverse features, even in later training stages.
For \textbf{mean Euclidean distances} between clusters,~\textsc{NoLimit} exhibits large distances between Epoch 1 and Epoch 5 but demonstrates minimal evolution between Epoch 5 and Epoch 10. 
This stagnation may highlight the model's failure to effectively generalize new rules. 
\textsc{DynamicLimit-Exp}, on the other hand, maintains substantial distances across epochs, indicating continuous embedding evolution and refinement throughout training.

\end{document}